\documentclass[conference]{IEEEtran}
\IEEEoverridecommandlockouts
\usepackage{cite}
\usepackage{amsmath,amssymb,amsfonts}
\usepackage{algorithmic}
\usepackage{graphicx}
\usepackage{textcomp}
\usepackage{xcolor}
\usepackage{multirow} 
\usepackage{url}
\usepackage{hyperref}
\def\BibTeX{{\rm B\kern-.05em{\sc i\kern-.025em b}\kern-.08em
    T\kern-.1667em\lower.7ex\hbox{E}\kern-.125emX}}
\begin{document}

\title{CBCTLiTS: A Synthetic, Paired CBCT/CT Dataset For Segmentation And Style Transfer}

\author{\IEEEauthorblockN{Maximilian E. Tschuchnig}
\IEEEauthorblockA{\textit{Salzburg University of Applied Sciences} \\
and \textit{University of Salzburg}\\
Salzburg, Austria\\
ORCID 0000-0002-1441-4752}
\and
\IEEEauthorblockN{Philipp Steininger}
\IEEEauthorblockA{\textit{MedPhoton GmbH}\\
\\
Salzburg, Austria}
\and
\IEEEauthorblockN{Michael Gadermayr}
\IEEEauthorblockA{\textit{Salzburg University of Applied Sciences} \\
\\
Salzburg, Austria\\
ORCID 0000-0003-1450-9222}
}

\maketitle

\begin{abstract}
Medical imaging is vital in computer assisted intervention. Particularly cone beam computed tomography (CBCT) with defacto real time and mobility capabilities plays an important role. However, CBCT images often suffer from artifacts, which pose challenges for accurate interpretation, motivating research in advanced algorithms for more effective use in clinical practice. 
   
In this work we present CBCTLiTS, a synthetically generated, labelled CBCT dataset for segmentation with paired and aligned, high quality computed tomography data. The CBCT data is provided in $5$ different levels of quality, reaching from a large number of projections with high visual quality and mild artifacts to a small number of projections with severe artifacts. This allows thorough investigations with the quality as a degree of freedom.
We also provide baselines for several possible research scenarios like uni- and multimodal segmentation, multitask learning and style transfer followed by segmentation of relatively simple, liver to complex liver tumor segmentation. 
CBCTLiTS is accesssible via \url{https://www.kaggle.com/datasets/maximiliantschuchnig/cbct-liver-and-liver-tumor-segmentation-train-data}.
\end{abstract}


\section{Introduction}

In the realm of computer-assisted interventions (CAI), precise and reliable imaging, especially intraoperative imaging, is paramount. Cone-Beam Computed Tomography (CBCT) is frequently employed to facilitate intraoperative interventions~\cite{rafferty2006intraoperative,roeder2023first} by providing detailed, three-dimensional representations of a patient's anatomy utilizing a cone-shaped X-ray beam and a flat-panel detector, integrated into a mobile system~\cite{jaffray2002flat}. However, rapid intraoperative imaging often comes with the disadvantage of significantly lower image quality~\cite{wei2024reduction} (compared to preoperative imaging), which can negatively impact the performance of downstream tasks such as segmentation. 

To assess the impact of reduced image quality on downstream tasks, the availability of image datasets with ground truth annotations is essential.
However, we identified a scarcity of publicly available CBCT datasets with ground truth annotations limiting the development and evaluation of automated CBCT image analysis methods. On the dataset sharing platform Kaggle, e.g. the search term \textit{CT Segmentation} returns $79$ datasets while the term \textit{CBCT Segmentation} returns only $1$. Due to this lack of publicly available datasets there is comparably little research in the field of CBCT image analysis, especially in the intraoperative setting. A comparison of Pubmed searches in the years $2014$ - $2024$ with the search strings \textit{(CT) AND (Segmentation) AND (intraoperative)} and \textit{(CBCT) AND (Segmentation) AND (intraoperative)} reveals, that research regarding intraoperative CBCT segmentation is only about $5.8\%$ as common as intraoperative computed tomography (CT) segmentation research.

\begin{figure*}[ht]
  \centering
  \mbox{} \hfill
  \includegraphics[width=\linewidth]{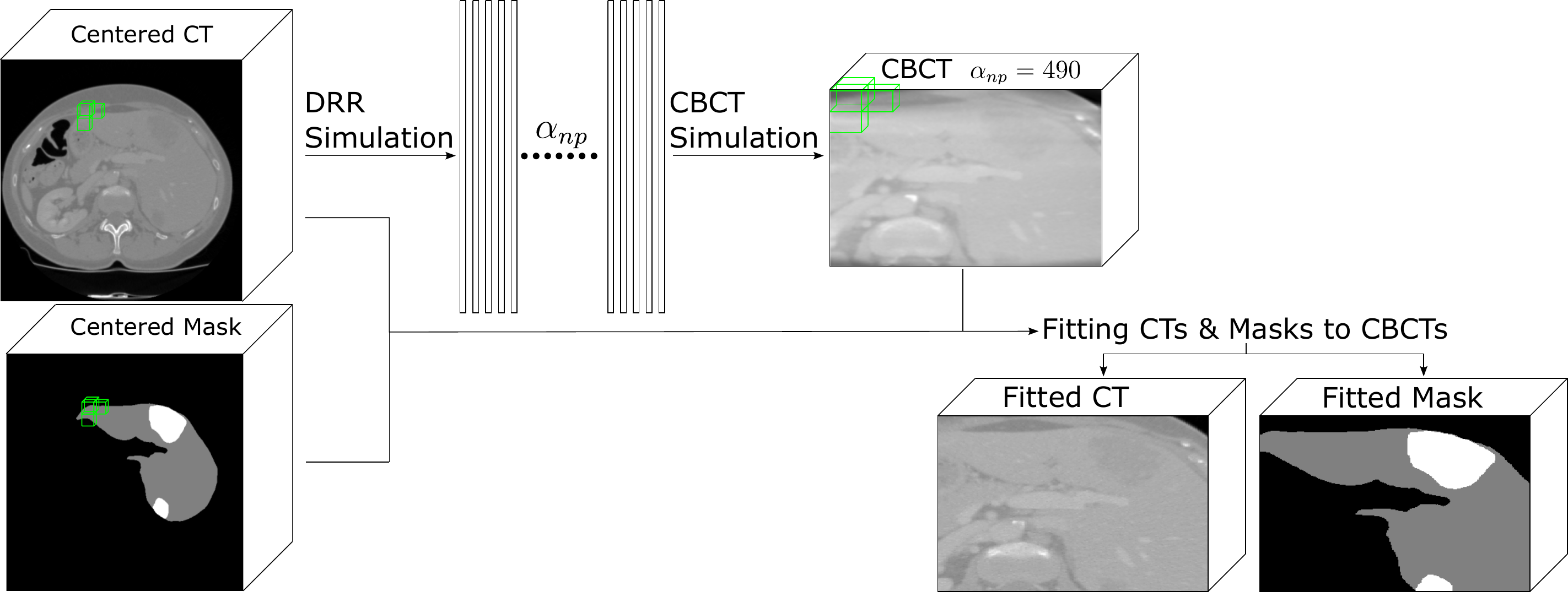}
  \hfill \mbox{}
  \caption{\label{fig:datapreprocessing}%
           Process of CBCTLiTS generation: after centering the original CT volumes around the liver (using the liver segmentations), DRRs were simulated from these centered CT. Then, CBCT were simulated and aligned with the original CT and masks.}
\end{figure*}

As stated by Bilc et al.~\cite{bilic2023liver}, the liver, as the largest solid organ in the human body, plays a crucial role in metabolism and digestion. It is also a common site for both primary and metastatic tumors, making it essential for comprehensive tumor staging and treatment planning. Primary liver cancer is the second leading cause of cancer-related deaths globally, and CT imaging is widely used to assess liver morphology, texture, and focal lesions. The Liver Tumor Segmentation Benchmark (LiTS)~\cite{bilic2023liver} dataset consists of overall $201$ abdominal volumes (with $194$ containing liver lesions), separated into a labelled training and an unlabelled testing dataset.
We chose this well studied dataset ($1008$ citations on Google Scholar) as a basis for our CBCT synthesised dataset due to the large amount of volumes, the availability of masks and the availability of two different targets corresponding to two different difficulty levels. These targets include the segmentation of relatively simple, regular and large liver areas and complex, small and irregular liver tumor areas.

\subsection{Contributions}

To enable thorough research in the field of CBCT analysis, a robust dataset with ground truth annotations is needed. We present CBCTLiTS, an adaptation of the LiTS dataset~\cite{bilic2023liver}, providing ground truth segmentation masks and paired CT samples. Our contributions include
\begin{itemize}
    \item Dataset Creation: Development of CBCTLiTS, including CBCT with varying volume quality ($5$ quality levels), ground truth segmentations and paired CT samples, facilitating various research scenarios such as unimodal and multimodal segmentation, multitask learning, and style transfer. The dataset consists of $201$ samples, $131$ in the training dataset with segmentations and $70$ in the testing dataset.
    \item Baseline Results: We provide baseline results for potential research directions, facilitated by CBCTLiTS, including unimodal and multimodal segmentation, multitask learning and style transfer.
\end{itemize}

\section{Dataset}

To generate the CBCTLiTS data, several steps, shown in Fig.~\ref{fig:datapreprocessing}, were taken. Initially, CT volumes were centered around either the liver (for training data with available liver masks) or the volumes middle (for test data). Following this, digitally reconstructed radiographs (DRR) were generated from the centered CT volumes, and utilized to synthesise CBCT scans. The CBCT volume quality was varied by adjusting the number of DRRs (undersampling) used in reconstruction. Finally, the original CT volumes and corresponding masks were fitted to the synthesised CBCT to match the same field-of-view.

As a basis for generating synthetic CBCT scans, real, diagnostic CT are needed. CT represents an imaging technology corresponding with a high visual quality with relatively few artifacts and improved image quality parameters like intensity homogeneity (compared to CBCT imaging). 
The LiTS CT dataset~\cite{bilic2023liver} was chosen as a basis to generate the synthetic CBCTLiTS data set. LiTS comprises $131$ abdominal CT scans in the training set and $70$ test volumes. The $131$ training volumes include segmentations of both the liver and liver tumors. This dataset contains data from seven different institutions with a diverse array of liver tumor pathologies, including primary and secondary liver tumors with varying lesion-to-background ratios. It also features a mix of pre- and post-therapy CT scans. The CT images were acquired using different scanners and acquisition protocols. For further details regarding the LiTS dataset, we refer to Bilic et al.~\cite{bilic2023liver}. LiTS was chosen due to its large number of samples ($201$), label balance (ranging from relatively simple, regular and large liver area to complex, small and irregular liver tumor area) and prominence with $1008$ citations (publication date $2023$, however, first results based on LiTS started in 2017). LiTS is also part of The Medical Segmentation Decathlon dataset~\cite{antonelli2022medical}.

\begin{figure*}[ht]
  \centering
  \mbox{} \hfill
  \includegraphics[width=\linewidth]{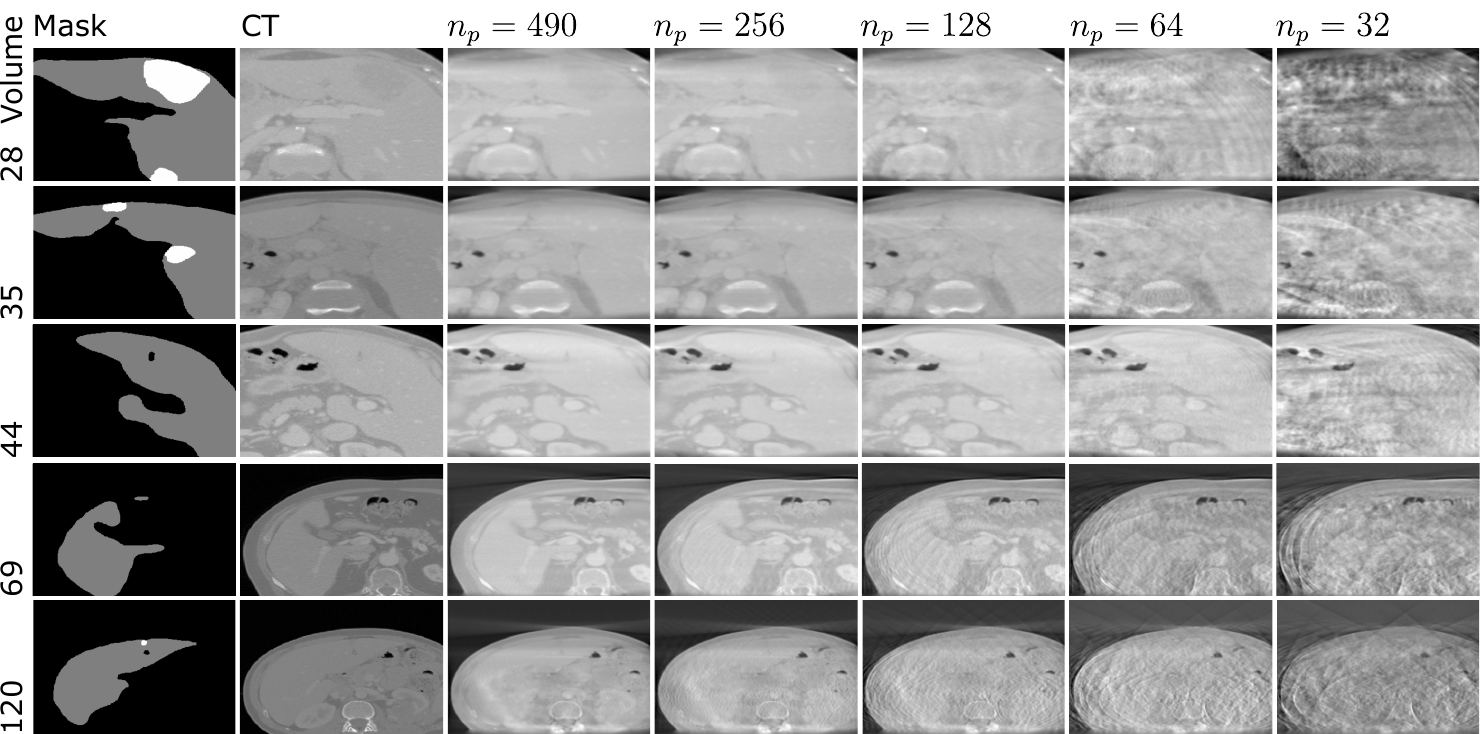}
  \hfill \mbox{}
  \caption{\label{fig:cbctlits}%
           Different volumes of CBCTLiTS data (vertically). Horizontally, the presented data consists of the aligned mask, the aligned CT and the $5$ different qualities of synthesised CBCT.}
\end{figure*}

In order to simulate CBCT from the CT volumes, DRRs were generated using Siddon ray tracing~\cite{siddon1985fast} under the assumption of a monochromatic X-ray spectrum. Given that standard CBCT with a centered detector typically exhibit small fields-of-view (approximately $25cm^3$), only specific areas of the original CT were extracted. To make effective use of the provided ground truth masks, the liver area was chosen as the center for CBCTLiTS training data. This center was calculated by calculating the center of gravity of the ground truth liver masks. The cone-beam projection geometry for DRR simulation was derived from the Loop-X (TM)~\cite{karius2022technical,karius2022quality,roeder2023first,wei2024reduction}, a mobile CBCT scanner used, e.g., in navigated spine surgery and radiation therapy. These projections were then used in a CBCT reconstruction algorithm, implementing filtered back-projection, specifically the Feldkamp-Davis-Kress (FDK)~\cite{feldkamp1984practical} algorithm. The reconstruction size of the CBCT was set to $252mm \times 246mm \times 250mm$ around the center with a voxel size of $0.688 \times 1.032 \times 0.688$. Unlike in the case of real-world CBCT scans, X-ray pre-processing steps such as beam hardening correction or Compton scatter correction were omitted, since idealized DRRs do not suffer from such effects but show primary signals only.

To simulate CBCT of different qualities, CBCT reconstruction was undersampled (by varying numbers of DRRs used for reconstruction). For different numbers of projections $n_p$, the angular gap between the generated DRRs was equidistantly distributed over a 360 degrees scan range of simulated full-fan acquisitions. The number of projections $n_p$ was varied between $\{490,256,128,64,32\}$ to generate scans with high visual quality similar to CT ($n_p$ = 490) as well as scans showing clear degradations ($n_p < 128$) and scans in between.
Tasks like segmentation and style transfer are facilitated by aligning the fields-of-view, rotation, and spacing of the CT and masks with the synthesized CBCT. To accomplish this, linear 3d transformation, combined with resampling is applied to align the mask and CT with the simulated CBCT.

\section{Experimental Setup}

To provide baselines for several possible research directions enabled by CBCTLiTS, this section introduces a segmentation baseline, based on unet~\cite{ronneberger2015u} and model adaptations investigating multitask learning (Subsect. \ref{subsec:mtl}), multimodal learning (Subsect. \ref{subsec:mml}) and style transfer (Subsect. \ref{subsec:sty}).
All evaluated segmentation models are based on 3D unet~\cite{cciccek20163d}. 
This 3D unet was adapted by applying batch norm after each layer in the double convolutional blocks. 

\begin{table*}[ht] 
\caption{This table shows results of several experiments, all based on CBCTLiTS. The first results show baseline segmentation scores using CBCT, both holistically and patched. Results are presented for different CBCT qualities and liver as well as liver tumor segmentation. Then, results of multitask learning approaches multitask-b and multitask-c, followed by multimodal learning approaches with different degrees of misalignment are presented. Finally, the basic 3D unet was applied to CBCT data that was style transfered into the CT domain. Data presented in bold was superior to its corresponding baseline (holistic or patched).} \label{tab:resultsAll}
\resizebox{\textwidth}{!}{\begin{tabular}{l|l|l|lllll|lllll}
& & &
\multicolumn{5}{c}{\textbf{Liver Segmentation}} & 
\multicolumn{5}{c}{\textbf{Liver Tumor Segmentation}} \\

& & &
\textbf{490} & 
\textbf{256} &
\textbf{128} &
\textbf{64} &
\textbf{32} &
\textbf{490} &
\textbf{256} &
\textbf{128} &
\textbf{64} & 
\textbf{32} \\ \hline



&
\textbf{base CBCT} & \cite{tschuchnig2024multi} &
0.884 
& 0.884
& 0.859
& 0.817
& 0.784
& 0.165
& 0.162
& 0.093
& 0.061 
& 0.029 \\ 

&
\textbf{base patch CBCT} & \cite{tschuchnig2024multi} &
0.811 
& 0.814
& 0.768
& 0.693
& 0.650
& 0.122
& 0.122
& 0.060
& 0.042 
& 0.028  \\ \hline \hline

\multirow{4}{*}{\ref{subsec:mtl}} &
\textbf{multitask-c} & \cite{tschuchnig2024multi} &
0.877 
& 0.879
& \textbf{0.874}
& \textbf{0.826}
& 0.771
& 0.149
& 0.153 
& 0.092 
& 0.058  
& \textbf{0.049}  \\ 

&
\textbf{multitask-b} & \cite{tschuchnig2024multi} & 
0.877  
& 0.884
& \textbf{0.871}
& \textbf{0.836}
& 0.775
& 0.149
& 0.147
& 0.078
& 0.051 
& 0.028  \\ 

&
\textbf{patch multitask-c} & \cite{tschuchnig2024multi} &
\textbf{0.847} 
& \textbf{0.821}
& \textbf{0.786}
& \textbf{0.720}
& 0.609
& \textbf{0.171}
& \textbf{0.127}
& \textbf{0.112}
& 0.036 
& \textbf{0.031} \\ 

&
\textbf{patch multitask-b} & \cite{tschuchnig2024multi} &
\textbf{0.847}  
& \textbf{0.831}
& \textbf{0.771}
& \textbf{0.720}
& 0.591
&  \textbf{0.171}
& \textbf{0.154} 
& \textbf{0.079} 
& 0.028  
& 0.021 \\ \hline \hline

\multirow{9}{*}{\ref{subsec:mml}} &
\textbf{perfect alignment} & \cite{tschuchnig2024multimodal} &
\textbf{0.933} 
& \textbf{0.931}
& \textbf{0.931}
& \textbf{0.933}
& \textbf{0.932}
& \textbf{0.330}
& \textbf{0.322}
& \textbf{0.298}
& \textbf{0.325} 
& \textbf{0.297} \\ \cline{2-13} 

&
\textbf{affine-s1} & \cite{tschuchnig2024multimodal} &
\textbf{\color{black}0.906\color{black}} & 
\textbf{\color{black}0.897\color{black}} & 
\textbf{\color{black}0.879\color{black}} & 
\textbf{\color{black}0.857\color{black}} & 
\textbf{\color{black}0.806\color{black}} &
\color{black}0.154\color{black} &  
\color{black}0.124\color{black} &  
\color{black}0.057\color{black} &  
\color{black}0.023\color{black} &  
\textbf{\color{black}0.051\color{black}} \\

&
\textbf{affine-s0.5} & \cite{tschuchnig2024multimodal} &
\textbf{\color{black}0.895\color{black}} & 
\textbf{\color{black}0.891\color{black}} & 
\textbf{\color{black}0.875\color{black}} & 
\textbf{\color{black}0.863\color{black}} & 
\textbf{\color{black}0.851} & 
\color{black}0.096 &  
\color{black}0.098 & 
\color{black}0.077\color{black} & 
\color{black}0.056\color{black} & 
\textbf{\color{black}0.107} \\

&
\textbf{affine-s0.25} & \cite{tschuchnig2024multimodal} &
\textbf{\color{black}0.904\color{black}} & 
\textbf{\color{black}0.906\color{black}} & 
\textbf{\color{black}0.893\color{black}} & 
\textbf{\color{black}0.883} & 
\textbf{\color{black}0.883} & 
\textbf{\color{black}0.181\color{black}} &  
\color{black}0.160\color{black} &  
\textbf{\color{black}0.163} &  
\textbf{\color{black}0.169} &  
\textbf{\color{black}0.171} \\

&
\textbf{affine-s0.125} & \cite{tschuchnig2024multimodal} &
\textbf{\color{black}0.908\color{black}} & 
\textbf{\color{black}0.900\color{black}} & 
\textbf{\color{black}0.889\color{black}} & 
\textbf{\color{black}0.884} & 
\textbf{\color{black}0.887} &  
\textbf{\color{black}0.187\color{black}} &  
\textbf{\color{black}0.166\color{black}} &  
\textbf{\color{black}0.166} &  
\textbf{\color{black}0.152} &  
\textbf{\color{black}0.185} \\ \cline{2-13} 

&
\textbf{elastic-s1} & \cite{tschuchnig2024multimodal} &
\textbf{\color{black}0.907\color{black}} & 
\textbf{\color{black}0.897\color{black}} & 
\textbf{\color{black}0.880\color{black}} & 
\textbf{\color{black}0.857\color{black}} & 
\textbf{\color{black}0.801\color{black}} &
\color{black}0.153\color{black} &  
\color{black}0.122\color{black} &  
\color{black}0.057\color{black} &  
\color{black}0.022\color{black} &  
\textbf{\color{black}0.049\color{black}} \\

&
\textbf{elastic-s0.5} & \cite{tschuchnig2024multimodal} &
\textbf{\color{black}0.895\color{black}} & 
\textbf{\color{black}0.891\color{black}} & 
\textbf{\color{black}0.870\color{black}} & 
\textbf{\color{black}0.845\color{black}} & 
\textbf{\color{black}0.800} & 
\color{black}0.098 &  
\color{black}0.103 &  
\color{black}0.085\color{black} &  
\color{black}0.029\color{black} &  
\textbf{\color{black}0.049} \\

&
\textbf{elastic-s0.25} & \cite{tschuchnig2024multimodal} &
\textbf{\color{black}0.893\color{black}} & 
\textbf{\color{black}0.898\color{black}} & 
\textbf{\color{black}0.869\color{black}} & 
\textbf{\color{black}0.853} & 
\textbf{\color{black}0.846} &
\color{black}0.154\color{black} &  
\color{black}0.142\color{black} &  
\textbf{\color{black}0.136} &  
\textbf{\color{black}0.129} &  
\textbf{\color{black}0.122} \\

&
\textbf{elastic-s0.125} & \cite{tschuchnig2024multimodal} &
\textbf{\color{black}0.910\color{black}} &  
\textbf{\color{black}0.901\color{black}} &  
\textbf{\color{black}0.891\color{black}} &  
\textbf{\color{black}0.885} &  
\textbf{\color{black}0.887} &  
\textbf{\color{black}0.186\color{black}} &  
\textbf{\color{black}0.168\color{black}} &  
\textbf{\color{black}0.163} &  
\textbf{\color{black}0.151} &  
\textbf{\color{black}0.168} \\ \hline \hline

\ref{subsec:sty} &
\textbf{sty-transfer CBCT} & & 
\textbf{0.901} 
& \textbf{0.888}
& \textbf{0.867}
& \textbf{0.835}
& \textbf{0.787}
& 0.104 
& 0.097 
& 0.065 
& 0.029  
& \textbf{0.040}  \\ \hline

\end{tabular}}
\end{table*}

All models were trained utilizing a sum of binary cross-entropy and Dice similarity. 
Further, the models were trained on an Ubuntu server using NVIDIA RTX A6000 graphics cards. To binarize the masks, a threshold of $0.5$ was applied to each channel of the unet (segmentation) output. All experiments were trained and evaluated $4$ times to facilitate stable results with the same random splits. Adam was used as an optimizer with a learning rate of $0.005$. For training the converted CBCTLiTS training data was separated into training-validation-testing splits containing ratios of $0.7$ (training), $0.2$ (validation), and $0.1$ (testing), respectively. In experiments where the whole volume was applied directly (holistic) the volumes had to be downscaled (isotropic) by the factor of two due to the large size of the volumes and memory restrictions ($48$ GB VRAM). 

\subsection{Combining segmentation and image reconstruction using multitask learning}
\label{subsec:mtl}

The aim of this adaptation is to investigate if the additional target of image reconstruction added to a segmentation model is beneficial for training the segmentation model~\cite{tschuchnig2024multi}. To accomplish this, a multistream architecture is utilized with two layers in the end, with the first layer establishing baseline segmentation. The second stream leads into a 3d convolutional layer with a linear activation function to facilitate image reconstruction. Therefore the updated loss contains a segmentation loss as well as an image reconstruction loss (l2). Two cases of this additional image reconstruction setting are investigated. The first investigated case is the reconstruction of the input CBCT (Table~\ref{tab:resultsAll}: multitask-c), i.e. the second stream is similar to an autoencoder. It aims to add morphological regularization to model training. The second case investigated how the reconstruction of the highest possible quality CBCT effects segmentation performance (Table~\ref{tab:resultsAll}: multitask-b) adding additional denoising effects. For more information we refer to Tschuchnig et al.~\cite{tschuchnig2024multi}.

\subsection{Multimodal learning, combining preoperative CT and intraopertative CBCT}
\label{subsec:mml}

To improve intraoperative imaging, one approach is to combine potentially low quality intraopterative volumes with potentially misaligned, high quality, preoperative volumes~\cite{tschuchnig2024multimodal}. This multimodal learning approach was investigated by performing early fusion of CT, as high quality, preoperative data and low quality CBCT, as intraoperative data. Further, different forms of misalignment with affine (random scaling, rotation and translation, Table~\ref{tab:resultsAll}: affine-s) and affine followed by elastic misalignment (Table~\ref{tab:resultsAll}: elastic-s) were performed, simulating real world scenarios with different degrees of alignment. To keep the number of parameters minimal, an alignment parameter $\alpha_a$ was introduced, controlling the maximum strength of all misalignment methods. Misalignment was performed using TorchIO RandomAffine and RandomElasticDeformation. For more information we refer to Tschuchnig et al.~\cite{tschuchnig2024multimodal}.

\subsection{Converting CBCT to CT using style transfer}
\label{subsec:sty}

Since the synthesised CBCT volumes are of lower visual quality then the CT volumes, a style transfer from the CBCT domain to the CT domain has the potential of segmentation performance increases. By performing this image to image style transfer, structureal CBCT artifacts and noise can theoretically be removed. This theory was experimentally investigated by using the CBCTLiTS test dataset to train a 3D Pix2Pix model, available at \url{https://github.com/MaxTschuchnig/TFVox2Vox}, for style transfer from the CBCT to the CT domain. Then, this model was applied to the CBCT training dataset to convert the CBCT scans of different data qualities to CT scans. The transformed data was subsequently used for liver and liver tumor segmentation using the baseline segmentation unet (Table~\ref{tab:resultsAll}: sty-transfer CBCT).


\section{Results}

Fig.~\ref{fig:cbctlits} shows example results of the data generation process with several CBCTLiTS samples. It displays, from left to right random samples of the masks and CT, aligned and in the same field of view as the generated CBCT followed by different visual quality levels of the CBCT. Liver lesions are labelled with the value $2$ (white) and the liver with the value $1$ (grey). The background voxels are labeled as $0$ (black).

Table~\ref{tab:resultsAll} shows results of different segmentation setups with the goals of liver and liver tumor segmentation. All presented results are based on CBCTLiTS, with the first results showing segmentation scores using CBCT, both holistically and patched. 
Further, results of $3$ different adaptations, multitask learning, multimodal learning and style transfer, using the unique components of CBCTLiTS are presented. All CBCT results are reported for the different CBCT qualities (labeled using the amount of projections $\alpha_{np}$) and for both liver and liver tumor segmentation. 
Baseline segmentation using the aligned, high quality CT reached a mean Dice of $0.930$ for liver and $0.303$ for liver tumor segmentation. For comparison, one of the best performing models of the original LiTS challenge reported scores of $0.962$ for liver and $0.739$ for liver tumor segmentation~\cite{isensee2021nnu,bilic2023liver}.

\section{Discussion}

The presented CBCTLiTS dataset is a unique dataset with paired CBCT and CT data. The dataset is also labelled, making it suitable for segmentation. In its most basic form, the CBCT dataset can be used to train segmentation models to segment relatively simple, regular and large liver areas and complex, small and irregular liver tumor areas. However, the different CBCT volume qualities enable a range of further research and evaluation possibilities, from simulating low dose CBCT to facilitating multitask learning (multitask-b). By far the most significant investigated segmentation performance increase resulted from the paired CBCT and CT data, used in the different multimodal learning approaches. Using multimodal learning, assuming perfect CT and CBCT alignment, even the baseline CT model could be beat in several cases. Here, the differing CBCT qualities again allow for important insights, showing a negative correlation between reducing volume quality and improvements attributable to multimodal learning. Finally, the pairing of CBCT and CT data allows for experimentation of image to image translation approaches. Here, further experimentation using e.g. diffusion models is enabled by CBCTLiTS.

In general, CBCTLiTS is a unique dataset useful for a multitude of research scenarios, especially for studies in the field of intraoperative medical imaging.

\section{Conclusion}

In this paper, we have introduced CBCTLiTS, a novel synthetic, paired CBCT/CT dataset designed to facilitate research in medical image segmentation and style transfer. CBCTLiTS addresses the scarcity of high-quality, publicly available CBCT datasets with annotated segmentations by providing ground truth segmentation masks and paired CT data. This dataset is available in five different CBCT quality levels, ranging from visual quality comparable to the original CT to low visual quality with a significant amount of artifacts.

Additionally to presenting the dataset and the process of dataset generation, several possible research scenarios, like uni and multimodal segmentation are introduced and baseline values for segmentation model adaptations are given, including holistic as well as patched approaches, These adaptations include multitask learning, multimodal learning and style transfer. Additional experiments show that CBCTLiTS is an especially capable resource to investigate intraoperative models. 

\subsection*{Acknowledgments}
This project was partly funded by the Austrian Research Promotion Agency (FFG) under the bridge project "CIRCUIT: Towards Comprehensive CBCT Imaging Pipelines for Real-time Acquisition, Analysis, Interaction and Visualization" (CIRCUIT), no. 41545455 and by the County of Salzburg under the project number 20102/F2300525-FPR (AIBIA).

\bibliographystyle{eg-alpha-doi} 
\bibliography{egbibsample}

\end{document}